\documentclass[review]{elsarticle}
\pdfoutput=1
\usepackage{lineno,hyperref}
\usepackage{graphicx}
\usepackage{bm}
\usepackage{amsmath,amsfonts}
\usepackage{multirow}
\usepackage{float}
\usepackage{subcaption}
\usepackage{endnotes}
\usepackage{natbib}
\usepackage{anysize}
\usepackage{xcolor}
\modulolinenumbers[5]
\newtheorem{theorem}{Theorem}
\newtheorem{proposition}{Proposition}

\journal{Computers \& Operations Research}

\usepackage{pdfpages}
\begin{document}
	\linespread{1.5}
	\includepdf{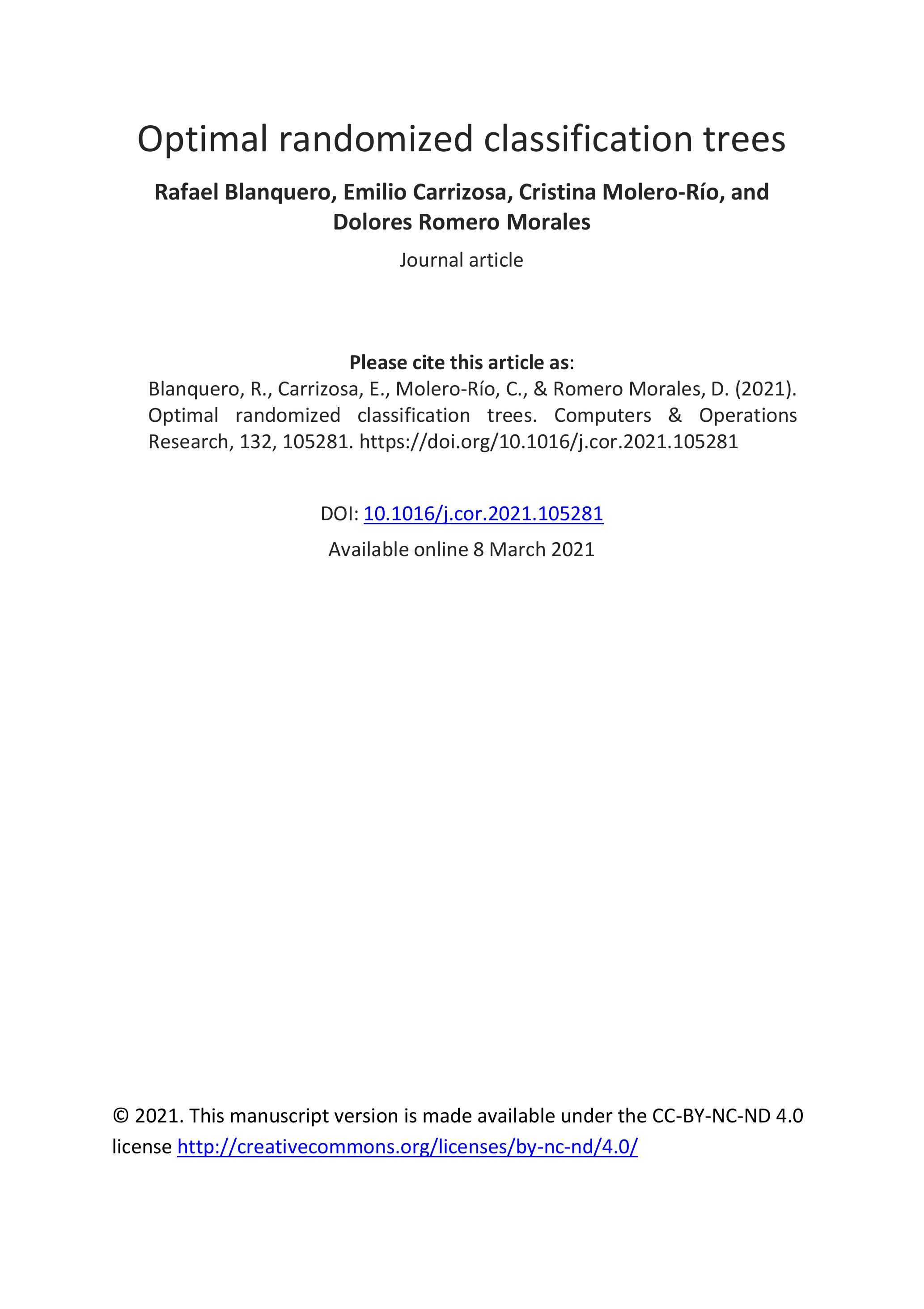}

\begin{frontmatter}

\title{Optimal Randomized Classification Trees}


\author[mymainaddress]{Rafael Blanquero}
\ead{rblanquero@us.es}

\author[mymainaddress]{Emilio Carrizosa}
\ead{ecarrizosa@us.es}

\author[mymainaddress]{Cristina Molero-R\'io\corref{mycorrespondingauthor}}
\cortext[mycorrespondingauthor]{Corresponding author}
\ead{mmolero@us.es}

\author[mysecondaryaddress]{Dolores Romero Morales}
\ead{drm.eco@cbs.dk}

\address[mymainaddress]{Instituto de Matem\'aticas de la Universidad de Sevilla (IMUS), Seville, Spain}
\address[mysecondaryaddress]{Copenhagen Business School (CBS), Frederiskberg, Denmark}

\begin{abstract}
Classification and Regression Trees (CARTs) are off-the-shelf techniques in modern Statistics and Machine Learning. CARTs are traditionally built by means of a greedy procedure, sequentially deciding the splitting predictor variable(s) and the associated threshold. This greedy approach trains trees very fast, but, by its nature, their classification accuracy may not be competitive against other state-of-the-art procedures. Moreover, controlling critical issues, such as the misclassification rates in each of the classes, is difficult. To address these shortcomings, optimal decision trees have been recently proposed in the literature, which use discrete decision variables to model the path each observation will follow in the tree. Instead, we propose a new approach based on continuous optimization. Our classifier can be seen as a randomized tree, since at each node of the decision tree a random decision is made. The computational experience reported demonstrates the good performance of our procedure.
\end{abstract}

\begin{keyword}
Classification and Regression Trees \sep Cost-sensitive Classification \sep Nonlinear Programming
\end{keyword}

\end{frontmatter}


\section{Introduction}
\label{intro}

 Extracting knowledge from data is a crucial task in Statistics and Machine Learning, which has applications in areas such as Biomedicine \citep{Jakaitiene2016,pardalosbiomed}, Criminal Justice \citep{jung2017simple,KleinbergQJE18,zengJRSC17}, Fraud Detection \citep{VlasselaerMS17}, Privacy Protection \citep{li2009against}, Health Care  \citep{bertsimasMS16,chaovalitwongse2008novel,souillardML16,ustunML16}, Risk Management \citep{BaesensMS03,martensEJOR07}, Social Networks \citep{fortunato2010community}, and Computational Optimization \citep{Khalil:2016:LBM:3015812.3015920}. Mathematical Optimization plays an important role in building such models and interpreting their output, see,  e.g.,  \cite{bertsimas2015or,bertsimas2014least,bertsimas2007classification,bottouSIAMR18,brooks2011support,carrizosa2018visualizing,carrizosa2017clustering,carrizosa2013supervised,fang2013right,FountoulakisMP2016,olafsson2008operations}.
	
	Decision trees constitute  a set of methods which can be considered as one of the most powerful tools for Classification and Regression in Statistics and Machine Learning \citep{hastie2009elements}. They are popular because they are rule-based and, when they are not very deep, deemed to be easy-to-interpret, see \cite{BaesensMS03,freitasACM14,goodman16,jung2017simple,ridgewayNIJJ13,ustunML16}.
	
	Since constructing optimal binary decision trees is known to be an NP-complete problem \citep{hyafil1976constructing}, researchers have traditionally focused on the design of heuristic procedures. CART \citep{breiman1984classification} is a popular algorithm for growing decision trees which, due to the complexity and the available computer technology at the time it was introduced, is based on a simple greedy and sequential partitioning procedure. In addition, CART, like many other tree algorithms \citep{yang2017regression}, uses orthogonal cuts, that is, each branching rule involves one single predictor variable. These characteristics of CART are aimed to achieve a low computational cost, but at the expense of accuracy. One of the most popular extensions of CARTs is Random Forests (RFs) \citep{biau2016random,breiman2001random,fernandez2014we,genuer2017random}, which involve building a collection of unpruned decision trees. In this way, better accuracies are typically obtained,  see e.g.\ \cite{fernandez2014we}, but having a complex and hardly interpretable decision structure. Nevertheless, there are ways to maintain the tree-like structure while improving accuracy. This includes combining several predictor variables in each split (oblique cuts), as well as choosing the splits globally along the tree, and not sequentially. These two strategies are at the core of this research.
	
Oblique cuts are more versatile than orthogonal cuts and tend to generate smaller trees, since several orthogonal cuts may be reduced to one single oblique cut. However, by their nature, oblique trees are harder to interpret and computationally more expensive than plain CARTs. { There are different algorithms that implement oblique cuts, such as OC1 \cite{murthy1994system} and \textit{oblique.tree} \cite{ripley2009fast}. OC1 is a greedy approach that finds oblique splits by means of a randomized perturbation algorithm. The methodology in \textit{oblique.tree} is also a greedy approach that is based on Mathematical Optimization: the oblique cuts are obtained as the output of a Logistic Regression classifier. A similar proposal can be found in \citep{orsenigo2003multivariate}, where the baseline classification procedure is a Support Vector Machine instead.} 
A different perspective dealing with oblique splits can be found in \cite{wang2015trading}, where the authors propose the so-called oblique tree sparse additive models. These are tree-structured probabilistic mixture-of-expert models, which are tackled using factorized asymptotic Bayesian inference, through EM-like iterative optimization.
	
	Due to the NP-completeness of the problem and the appeal of small trees to ease interpretability, optimal trees are grown up to a given maximum depth level. Two bottom-up approaches for finding optimal trees with the smallest classification error within the class of all possible trees of a specified depth are outlined in \cite{savicky2000optimal}, being exponentially complex in the number of predictor variables. In \cite{bennett1996optimal,norouzi2015efficient}, the construction of optimal oblique trees in a single step (non-greedily) is considered. In \cite{bennett1996optimal}, the topology of the tree is fixed, including the assignment of classes to leaf nodes, and represents it as a set of disjunctive linear inequalities. This yields a non-linear non-convex continuous optimization problem over a polyhedral region. Since the local search (performed by Frank-Wolfe algorithm) may get stuck at local optima,
	a heuristic search approach, the so-called Extreme Points Tabu Search, is also proposed. Standard greedily built trees are outperformed 
	by this approach for medium-size two-class problems. A method for non-greedy learning of oblique trees is developed in \cite{norouzi2015efficient}, in which a convex-concave upper bound on the tree's empirical loss  is optimized using Stochastic Gradient Descent, enabling effective training with larger data sets.
	
	Some attempts on bi-criteria optimization of decision trees have been proposed recently. For instance, \cite{chikalov2018bi} describes algorithms which allow to construct the set of Pareto optimal points for a given decision table and different objective functions, modeling the size of the tree, by controlling the depth and number of nodes, among others. One of its limitations is that it only handles categorical predictor variables and works efficiently only with medium-size decision tables. 
	
	Constructing an optimal decision tree involves, in principle, many discrete decisions, such as deciding whether splitting a branch node or indicating the path in the tree from the root node to the leaf node that each observation will follow. This is the approach followed  in recent papers such as
	\cite{bertsimas2017optimal,menickelly2016optimal,verwer2017learning,verwer2017auction}, which consider mixed-integer formulations. While these proposals are deterministic, our approach is a randomized one, where a random decision is made at each non-terminal node of the tree with a certain probability. Our classifier, the Optimal Randomized Classification Tree (ORCT), can be built solving a continuous optimization formulation. There are other important similarities/differences between our method and those mentioned above. First, our randomized methodology naturally provides probabilistic output on class membership tailored to each individual, in contrast to existing approaches, where all individuals in the same leaf node have associated the same probability. Second, in terms of predictor variables, and as in \cite{bertsimas2017optimal}, we deal with any type of predictor variables, linearly scaling them to the 0-1 interval and creating dummies when needed. In \cite{verwer2017learning,verwer2017auction}, they propose a non-linear scaling that assigns a unique integer to each value of each predictor variable, maintaining the original order of those values. Among others, one of the benefits of
	doing this scaling is that thresholds in branch nodes will be also represented by integers and therefore mixed-integer programming solvers will find it easier to branch on these values. The approach in \cite{menickelly2016optimal} focuses on categorical predictor variables by exploiting the resulting combinatorial structure of considering every possible subset of categories of a given predictor variable. However, when there is a predictor variable with many categories, the number of resulting 0-1 predictor variables explodes. Third, in terms of type of trees, orthogonal ones are grown in \cite{bertsimas2017optimal,menickelly2016optimal}. The former also constructs oblique trees, as we do. In \cite{verwer2017learning,verwer2017auction}, the authors construct oblique trees where the 0--1 coefficients in orthogonal cuts are generalized to integers. Fourth, in terms of the number of classes, multiclass problems can be handled with any approach, except for \cite{menickelly2016optimal}, which is restricted to two classes. Fifth, in relation to the topology of the tree, \cite{bertsimas2017optimal,verwer2017learning,verwer2017auction} specify its depth, and penalize the size of the tree in the objective function so that trees may be smaller than the pre-established depth. In \cite{menickelly2016optimal}, the authors specify a priori every leaf node in the tree and, as in our approach, a maximal tree for a given depth is constructed. Sixth, the feasible region in our formulation, that is, the number of decision variables as well as the number of constraints, is independent of the training sample size $N$, and, therefore, our approach scales up when $N$ grows. Finally, we also provide, as in \cite{menickelly2016optimal,verwer2017learning,verwer2017auction}, the flexibility we can borrow from Mathematical Optimization \citep{carrizosa2013supervised}, by controlling the classification performance in those classes where misclassification errors are more critical.
	
	From the computational perspective, a serious drawback of existing optimization-based procedures to build trees is their running times. This is particularly critical in integer programming-based strategies, for which the running times may dramatically increase  with the data dimensionality. These limitations are discussed in \cite{dunn2018optimal}, where a local-search heuristic is developed which is based on the integer programming approach in \cite{bertsimas2017optimal}. This heuristic allows the user to build deeper trees that generally outperform those in \cite{bertsimas2017optimal}. Using, as we do, a continuous optimization-based method, yields typically, as reported in Section \ref{Computational experiments}, better results in a more reasonable time than  other combinatorial optimization-based approaches as those in \cite{bertsimas2017optimal}.

	In summary, the main contributions of this paper are outlined below:
	\begin{enumerate}
		\item A novel continuous-based approach for building optimal classification trees is provided. The randomization of our approach as well as the inclusion of oblique cuts allow the removal of the integer decision variables present in recent proposals in the literature. The feasible region in our formulation is independent of the size of the training sample $N$. Therefore, our approach scales up when $N$ grows.
		\item Our approach naturally provides probabilistic output on class membership tailored to each individual, in contrast to existing approaches, where all individuals in the same leaf node are assigned the same probability.
		\item Preferences on classification rates in critical classes are successfully handled, unlike heuristic approaches like CART, { OC1, \textit{oblique.tree}} or RF, which can not address this issue explicitly.
		\item Our numerical results illustrate the outperformance of our approach in terms of accuracy against CART { and OC1} as well as recent proposals based on integer programming as those in \cite{bertsimas2017optimal}. Moreover, we { are comparable to \textit{oblique.tree} and} manage to get close to improved methods like RFs and the local-search heuristic in \cite{dunn2018optimal}, respectively. 
	\end{enumerate}
	
	The remainder of the paper is organized as follows. In Section \ref{Optimal Randomized Classification Trees}, we introduce the ORCT and its formulation, and show some theoretical properties. In Section \ref{Variants of ORCTs}, we introduce some variants of ORCTs, including the possibility to control the classification performance in critical classes. In Section \ref{Computational experiments}, computational experiments with ORCTs are reported. The results obtained are compared with CART, {OC1, \textit{oblique.tree},}  the integer programming-based approach in \cite{bertsimas2017optimal} and its related local-search heuristic in \cite{dunn2018optimal}, and RFs. Finally, conclusions and possible lines of future research are provided in Section \ref{Conclusions and future research}.

\section{Optimal Randomized Classification Trees}
\label{Optimal Randomized Classification Trees}
\subsection{The basic idea}
\label{The basic idea}

Suppose we are given a training sample of $N$ individuals, $I=\left\lbrace\left(\bm{x}_i,y_i\right)\right\rbrace_{ 1\leq i \leq N}$, on which $p$ predictor variables are measured, $\bm{x}_i\in\mathbb{R}^p$, and one class label is associated with each one, $y_i\in\left\lbrace 1,\ldots, K\right\rbrace$. Let $I_k\subset I$ denote the set of individuals in $I$ in class $k$, and $|I_k|$ its cardinality. For simplicity, numerical predictor variables are considered. Without loss of generality, we assume that $\bm{x}_i\in\left[0,1\right]^p$.

Classic decision tree methods consist of sequentially, and greedily, partitioning the predictor space $\left[0,1\right]^p$ into disjoint sets or nodes by imposing certain conditions on the predictor variables. The usual splitting criterion is to take the split that makes descendant nodes purer, i.e., nodes with observations more and more homogenous in terms of class membership. The process of partitioning finishes when a stopping criterion is satisfied. Then, leaf nodes are labeled with a class label, $1,\ldots,K$. Commonly, a leaf node is labeled with the most frequent class in the individuals that have fallen into the node. Once the tree is built, the prediction of future unlabeled data is done in a deterministic way. Given a new observation $\bm{x}$, starting from the root node, it will end up in a leaf node, depending on the values the predictor variables take, and its predicted class will be the class label of that leaf node. Alternatively, the prediction of future unlabeled data can be done in a probabilistic way, using the relative frequencies of belonging to each class of the corresponding leaf node.

The approach proposed here is different: prediction is randomized. At each node of an Optimal Randomized Classification Tree (ORCT), a random variable will be generated to indicate by which branch one has to continue. Since we are building binary trees, the Bernoulli distribution is appropriate, whose probability of success will be determined by the value of a cumulative density function (CDF), evaluated over the vector of predictor variables. In this way, each leaf node will not contain a subset of individuals but all the individuals in the training sample, for which the probability of falling into such leaf node is known. Finally, the class label of each leaf node will be a decision variable, which will be found by minimizing the expected misclassification cost over the whole training sample.

\begin{figure}[H]
	\centering
	\includegraphics[scale=1]{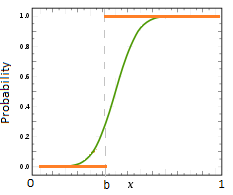}
	\caption{ The probability of an individual going down by the right branch is depicted for both types of trees: the classic approach (orange line) and the proposed ORCT (green line).}
	\label{logistica}
\end{figure}

The distinctive element in our approach is the fact that the yes/no rule in decision tree methods is replaced by a soft rule, induced by a continuous 
CDF. In this way, a smoother and therefore more stable rule is obtained. Indeed, suppose that the first cut of a classic classification tree forces individuals with  $X\leq b$  to go down the tree by the left branch. Then, for an incoming individual with $X=b+\varepsilon$, $\varepsilon>0$ sufficiently small, classic decision tree methods would give them probability of going down the branch $X > b$ equal to $1$ and $0$, otherwise, as in  the orange line in \figurename{ \ref{logistica}}. To avoid the discontinuities present in tree-based estimators like CART and RF, we propose to smooth the probabilities of going left or right in a neighbourhood of $b$ through a CDF, see the green line in \figurename{ \ref{logistica}}.

\subsection{The model}

After having introduced how ORCTs work, we will next formulate the problem.

\begin{figure}[H]
	\centering
	\includegraphics[scale=0.45]{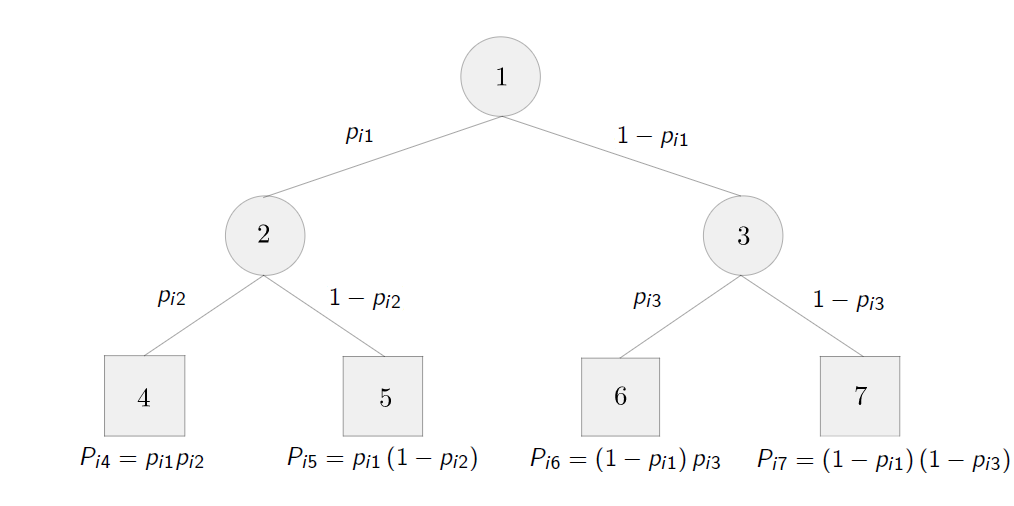}
	\caption{Optimal Randomized Classification Tree for depth $D=2$.}
	\label{Optimal randomized classification tree for depth $D=2$.}
\end{figure}

Our approach starts from the maximal binary tree of depth $D$, i.e.,  a binary tree in which each branch node has two children and terminal nodes all have the same depth, namely, $D$. For instance, \figurename{ \ref{Optimal randomized classification tree for depth $D=2$.} shows the maximal binary tree of depth $D=2$. Given $D$, the total number of nodes is known in advance, $T=2^{\left(D+1\right)}-1$. The sets of branch and leaf nodes are known and numbered as follows:
	
	Branch nodes: nodes $t \in \tau_B = \left\lbrace 1,\dots,\lfloor T/2 \rfloor \right\rbrace$.
	
	Leaf nodes: nodes $t \in \tau_L = \left\lbrace \lfloor T/2 \rfloor +1,\ldots,T\right\rbrace $.
	
	Oblique splits are modeled through linear combinations of the predictor variables. To do that, we need to define, for each $j=1,\ldots,p$ and each $t \in \tau_B$, the decision variables $a_{jt} \in \left[-1,1\right]$ to indicate the value of the coefficient of predictor variable $j$ in the oblique cut over branch node $t\in\tau_B$. The $p\times\vert \tau_B\vert$ matrix of these coefficients will be denoted by $\bm{a} = \left(a_{jt}\right)_{j=1,\ldots, p, t\in\tau_B}$, and the expressions $\bm{a}_{j\cdot}$ and $\bm{a}_{\cdot t}$ will stand for the $j$-th row and the $t$-th column of $\bm{a}$, respectively.
	The intercepts of the linear combinations correspond to decision variables $\mu_t \in \left[-1,1\right]$, seen as the location parameter at every branch node $t\in\tau_B$. Let $\bm{\mu}$ be the vector that comprises every $\mu_t$, i.e., $\bm{\mu}= \left( \mu_t\right)_{t\in\tau_B}$.
	
	Now, a univariate continuous CDF $F\left( \cdot \right)$ centered at $0$ is assumed. Then, for each individual $i=1,\ldots,N$ at each branch node $t\in\tau_B$, the parameter of their corresponding Bernoulli distribution is obtained as follows:
	\begin{equation}
	\label{cprima3} p_{it}\left(\bm{a}_{\cdot t},\mu_t\right)= F\left(\dfrac{1}{p} \sum_{j=1}^p a_{jt}x_{ij} -\mu_t\right), \,\, i =1,\ldots, N, \,\, t\in \tau_B.
	\end{equation}
	Note that this probability is a continuous function in the predictor variables $\bm{x}_i$, since the CDF $F$ is a continuous function.
	
	The value $p_{it}\left(\bm{a}_{\cdot t},\mu_t\right)$ will be used in the corresponding left branch and $1-p_{it}\left(\bm{a}_{\cdot t},\mu_t\right)$ in the right one, as seen in \figurename{ \ref{Optimal randomized classification tree for depth $D=2$.}}. We denote as $N_L(t)$ the set of ancestor nodes of node $t$ whose left branch takes part in the path from the root node to node $t$. Respectively, $N_R(t)$ is the set of ancestor nodes of node $t$ whose right branch takes part in the path from the root node to $t$. If $N(t)$ denotes the set of ancestors of node $t$, we have that $N(t)=N_L(t)\cup N_R(t)$. For leaf node $t=5$ in \figurename{ \ref{Optimal randomized classification tree for depth $D=2$.}}: $N_L(5)=\left\lbrace 1\right\rbrace$, $N_R(5)=\left\lbrace 2 \right\rbrace$ and $N(5)=\left\lbrace 1,2\right\rbrace$.
	
	Once these sets are defined, we can obtain the probability of an individual falling into a given leaf node:
	\begin{equation}
	\label{c2}
	P_{it}\left(\bm{a},\bm{\mu}\right)\equiv  \mathbb{P}\left(\bm{x}_i\in t\right)= \prod_{t_l\in N_L(t)} p_{it_l}\left(\bm{a}_{\cdot t_{l}},\mu_{t_{l}}\right)\prod_{t_r\in N_R(t)}\left(1- p_{it_r}\left(\bm{a}_{\cdot t_{r}},\mu_{t_{r}}\right)\right), \,\, i=1,\ldots,N,\,\, t\in \tau_L.
	\end{equation}
	As a consequence of \eqref{cprima3}, this probability is a continuous function in the predictor variables $\bm{x}_i$.
	
	Now, it is necessary to define, for each leaf node $t\in \tau_L$, the binary decision variables $\bm{C}=(C_{k t})_{k=1,\ldots,K,t\in \tau_L}$ that model the class label assigned to each of them, where \begin{equation}
	\nonumber
	C_{kt}= \left\lbrace
	\begin{array}{ll}
	1, & \text{if node $t$ is labeled with class $k$}\\
	0, & \text{otherwise}
	\end{array}\right., k=1,\ldots,K,\,\, t\in \tau_L.
	\end{equation}
	
	We must add the following set of constraints for making a single class prediction at each leaf node:
	\begin{equation}
	\nonumber
	\sum_{k=1}^K C_{kt} = 1, \,\, t\in \tau_L.
	\end{equation}
	
	As a natural strengthening, we force each class $k=1,\ldots, K$ to be identified by, at least, one terminal node, by adding the set of constraints below:
	\begin{equation}
	\label{strenght}
	\sum_{t\in \tau_L} C_{kt} \geq 1, \,\, k=1,\ldots, K,
	\end{equation} where it is implicitly assumed  that $K\le 2^D$ so that the previous constraints make sense. This set of constraints prevent the observations belonging to the minority classes from being fully misclassified. Nevertheless, they could be easily removed when desired.
	
	For fixed $\bm{a},\bm{\mu},\bm{C}$, the probability of individual $i$ being assigned to class $k$ is equal to 
	\begin{equation}
	\label{classmembershipprobabilities}
	\sum_{t\in\tau_L} P_{it}\left(\bm{a},\bm{\mu}\right) C_{kt}.
	\end{equation}
	Using the continuity of $P_{it}\left(\bm{a},\bm{\mu}\right)$ in the predictor variables $\bm{x}_i$, we have that small changes in $\bm{x}_i$ lead to small changes in the values of the probabilities of class membership in \eqref{classmembershipprobabilities}. We illustrate this amenable property of our approach using the balanced two-class simulated data set in Figure \ref{Simulated data set with $p=2$, $K=2$ and $N=400$ to compare the probabilities of class membership derived from ORCT with those derived from CART (deterministic and probabilistic) and RF.}. 	\begin{figure}
		\centering
		\includegraphics[scale=0.6]{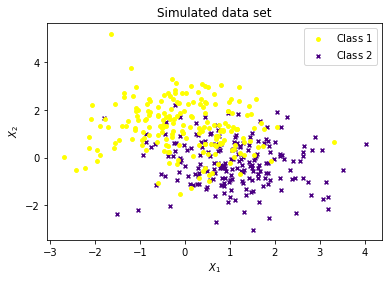}
		\caption{Simulated data set with $p=2$, $K=2$ and $N=400$ to compare the probabilities of class membership derived from ORCT with those derived from CART (deterministic and probabilistic) and RF.}
		\label{Simulated data set with $p=2$, $K=2$ and $N=400$ to compare the probabilities of class membership derived from ORCT with those derived from CART (deterministic and probabilistic) and RF.}
	\end{figure}The data set consists of $N=400$ individuals equally split into the two classes, characterized by $p = 2$ predictor variables. The predictor variables for individuals labeled as class $k$, $k=1,2$, have been generated following a bivariate normal distribution, $\mathcal{N}\left(\bm{\eta}_k,\bm{\Sigma}_k\right)$. We have chosen $\bm{\eta}_1=\left(0.00,1.25\right)^\top$, $\bm{\eta}_2=\left(1.00,-0.25\right)^\top$ and $\bm{\Sigma}_1=\bm{\Sigma}_2$ the identity matrix of size 2.
	In Figure \ref{Heatmap of probabilities of class membership for deterministic CART, probabilistic CART, RF and ORCT on the simulated data set}, 	\begin{figure}
		\centering
		\includegraphics[scale=0.6]{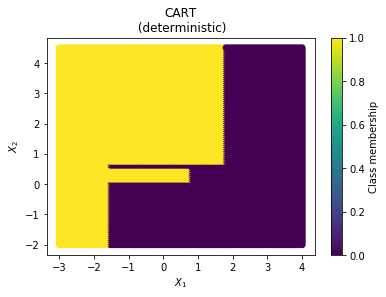}\includegraphics[scale=0.6]{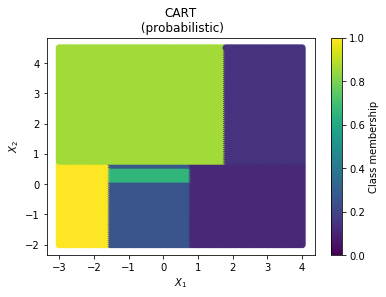}
		\includegraphics[scale=0.6]{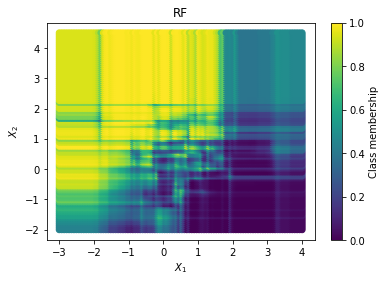}\includegraphics[scale=0.6]{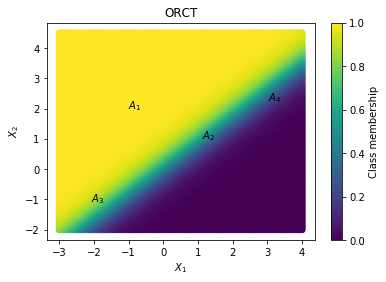}
		\caption{Heatmap of probabilities of class membership for deterministic CART, probabilistic CART, RF and ORCT on the simulated data set in Figure \ref{Simulated data set with $p=2$, $K=2$ and $N=400$ to compare the probabilities of class membership derived from ORCT with those derived from CART (deterministic and probabilistic) and RF.}.}\label{Heatmap of probabilities of class membership for deterministic CART, probabilistic CART, RF and ORCT on the simulated data set}
	\end{figure} we compare the probabilistic output of ORCT with the one derived from the two versions of CART described in Subsection \ref{The basic idea} (a deterministic as well as a probabilistic one) and RF. The CART classifier is built with the \texttt{rpart} R package \citep{rpartR} and RF with the \texttt{randomForest} R package \citep{rfR} both with the default tuning parameters, while the ORCT classifier outlined below is built with depth $D=1$. For each classifier, we derive the probability of belonging to class $k=1$ and use a heatmap plot to visualize it. Clearly, our ORCT approach is able to produce smoother class membership probabilities. We illustrate these probabilities at four different points, namely, $A_1=\left(-1.0,2.0\right)$, $A_2=\left(1.1,1.0\right)$, $A_3=\left(-2.1,-1.1\right)$ and $A_4 = \left( 3.0,2.3\right) $. On the yellow zone, the probability of belonging to class $k = 1$ for $A_1$ is equal $0.9998$. If one is placed on the oblique cut, almost no-discriminatory probabilities are obtained. Indeed, the probability of belonging to class $k = 1$ for $A_2$ is equal to $0.5077$. Above the cut, the probability of belonging to class $k=1$ increases smoothly. See $A_3$ on the green zone, for which the probability of belonging to class $k = 1$ is equal to $0.8701$. Likewise, below the cut, the probability of belonging to class $k=1$ decreases instead. For  $A_4$, this probability is equal to $0.2438$.

	Once the probabilities in \eqref{classmembershipprobabilities} have been defined, we can now model the objective function of our model. As said before, the objective is to minimize the expected misclassification cost over the sample, so we need to introduce a misclassification cost for classifying an individual $i$, whose class is $y_i$, in class $k$:
	\begin{equation}
	\label{w}
	W_{y_ik} \geq 0,\,\, k=1,\ldots,K.\\
	\end{equation} We define $W_{y_ik} = 0$ if $y_i=k,\,\, k=1,\ldots,K$.
	
	Thus, the objective function takes the following form: 
	\begin{equation}
	\label{objective2}
	\dfrac{1}{N}\sum_{i=1}^N \sum_{t\in\tau_L} P_{it}\left(\bm{a},\bm{\mu}\right) \sum_{k=1}^K W_{y_ik}C_{kt}.
	\end{equation} 
	Thus, given the sample $I$ split into $K$ classes, the CDF $F$, the depth of the tree $D$ and the misclassification costs $W_{y_ik}$, a mixed-integer non-linear optimization (MINLO) problem to build the proposed classification tree reads as follows:
	\begin{align}
	\begin{split}
	{\min} \hspace*{0.6cm}&   \dfrac{1}{N}\sum_{i=1}^N \sum_{t\in\tau_L} P_{it}\left(\bm{a},\bm{\mu}\right) \sum_{k=1}^K W_{y_ik}C_{kt}\\
	\text{s.t. \hspace*{0.5cm}} &  \sum_{k=1}^K C_{kt} = 1, \,\, t\in \tau_L,\\
	& \sum_{t\in \tau_L} C_{kt} \geq 1, \,\, k=1,\ldots, K, \\
	& a_{jt} \in \left[-1,1\right],\,\, j=1,\ldots,p,\,\, t\in \tau_B,\\
	& \mu_{t} \in \left[-1,1\right],\,\, t\in \tau_B,\\
	& C_{kt} \in \left\lbrace 0,1\right\rbrace, \,\, k=1,\ldots,K,\,\, t\in \tau_L,\label{MINLO}
	\end{split}
	\end{align} where $P_{it}\left(\bm{a},\bm{\mu}\right)$ is defined as in \eqref{c2}.
	
	The presence of binary decision variables in a framework where the objective function is highly complex non-convex could appear to be discouraging. Nevertheless, without loss of optimality, we can relax the binary decision variables $C_{kt},\,\, k=1,\ldots,K,\,\, t\in\tau_L$,
	 yielding to the continuous formulation we were looking for. Theorem \ref{transportationproblem} guarantees the equivalence of the resulting Non-Linear Continuous Optimization (NLCO) problem and the MINLO problem.
	
	The NLCO problem, which will be referred henceforth as the Optimal Randomized Classification Tree (ORCT), reads as follows:
	\begin{align}
	\begin{split}
	{\min} \hspace*{0.6cm}&   \dfrac{1}{N}\sum_{i=1}^N \sum_{t\in\tau_L} P_{it}\left(\bm{a},\bm{\mu}\right) \sum_{k=1}^K W_{y_ik}C_{kt}\\
	\text{s.t. \hspace*{0.5cm}} &  \sum_{k=1}^K C_{kt} = 1, \,\, t\in \tau_L,\\
	& \sum_{t\in \tau_L} C_{kt} \geq 1, \,\, k=1,\ldots, K, \\
	& a_{jt} \in \left[-1,1\right],\,\, j=1,\ldots,p,\,\, t\in \tau_B,\\
	& \mu_{t} \in \left[-1,1\right],\,\, t\in \tau_B,\\
	& { C_{kt} \geq 0}, \,\, k=1,\ldots,K,\,\, t\in \tau_L,\label{NLCO}
	\end{split}
	\end{align} where $C_{kt},\,\,k=1,\ldots,K,\,\, t\in\tau_L$, can be seen as the probability that the class label $k$ is assigned to leaf node $t$. 

	\begin{theorem}
		\label{transportationproblem}
		There exists an optimal solution to \eqref{NLCO} such that $C_{kt}\in\left\lbrace 0,1\right\rbrace,\,\, k=1,\ldots,K,\,\,t\in\tau_L$.
	\end{theorem}
	
	\noindent \textit{Proof of Theorem \ref{transportationproblem}}. The continuity of the objective function, defined in \eqref{NLCO} over a compact set, ensures the existence of an optimal solution of the optimization problem, by the Weierstrass Theorem. Let $(\bm{a}^*,\bm{\mu}^*,\bm{C}^*)$ be an optimal solution to \eqref{NLCO}. Fixing $(\bm{a}^*,\bm{\mu}^*)$, we have the following problem on the decision variables $C_{kt},\,\,k=1,\ldots,K,\,\,t\in\tau_L$:
	\begin{align}
	\begin{split}
	\nonumber
	{\min} \hspace*{0.6cm}&   \dfrac{1}{N}\sum_{i=1}^N \sum_{t\in\tau_L} P_{it}\left(\bm{a}^*,\bm{\mu}^*\right) \sum_{k=1}^K W_{y_ik}C_{kt}\\
	\text{s.t. \hspace*{0.5cm}} & \sum_{k=1}^K C_{kt} = 1, \,\, t\in \tau_L,\\
	& \sum_{t\in \tau_L} C_{kt} \geq 1, \,\, k=1,\ldots, K, \\
	& { C_{kt} \geq 0}, \,\, k=1,\ldots,K,\,\, t\in \tau_L,
	\end{split}
	\end{align} a transportation problem for which the existence of an integer optimal solution is well-known to hold, i.e., there is $\bm{\overline{C}}=(\overline{C}_{kt})$, with $\overline{C}_{kt} \in \{0,1\},\,\, k=1,\ldots,K,\,\, t\in \tau_L,$ such that $(\bm{a}^*,\bm{\mu}^*,\bm{\overline{C}})$ is also optimal for \eqref{NLCO}. \hfill $\square$\\
	
	The prediction of future unlabeled data with predictor variables $\bm{x}$ that ORCT makes is probabilistic by construction, namely, the probabilities in \eqref{classmembershipprobabilities} are returned where $\bm{x}_i$ is replaced by $\bm{x}$. In our computational experience, this prediction is made in a deterministic fashion by choosing the class for which the class membership probability is the highest.
	
	\subsection{Theoretical properties}
	
	In this section, we explore some theoretical properties of our approach. First, we show the relationship between ORCT, which uses a randomized decision rule in each branch node, and other optimization-based approaches, which are deterministic, as CARTs are. We will prove that when the level of randomization decreases to zero, our ORCT converges to an Optimal Deterministic Classification Tree (hereafter, ODCT), i.e., those in \cite{bertsimas2017optimal,verwer2017learning,verwer2017auction}. Second, we prove asymptotic results for the optimization problem attached to our ORCT when the training sample size grows to infinity. 
	
	We start by investigating the relationship between ORCT and the so-called ODCTs. Recall that ORCT uses the CDF $F$ to make, at each branch node, the decision to move to the left or to the right child node, while ODCT makes this decision in a deterministic fashion. In order to show the convergence of ORCT to ODCT, we define a family of CDFs $F_\gamma$, parametrized by $\gamma>0$, such that
	\begin{equation}
	\label{Fgamma}
	\lim_{\gamma\to \infty} F_\gamma\left(\cdot\right)= \left\lbrace
	\begin{array}{ll}
	1, & \text{if $\left(\cdot\right)\geq 0$}\\
	0, & \text{otherwise}
	\end{array}\right..
	\end{equation} An example of this family can be defined using the logistic CDF as follows 
	\begin{equation}
	\label{logisticCDF}
	F_\gamma\left(\cdot\right) = \dfrac{1}{1+\exp\left( -\left( \cdot\right)\gamma\right)},\,\, \gamma>0,
	\end{equation} provided that the argument is different from zero.
	
	For each value of the parameter $\gamma$ we have an ORCT, say ORCT($\gamma$). The larger the value of $\gamma$, the closer the decision rule defined by $F_\gamma$ is to a deterministic rule. In the limit case, when $\gamma$ is equal to $\infty$, the decision rule is deterministic, using \eqref{Fgamma}, and therefore the corresponding optimal classification tree is an ODCT. Thus, it is now easy to show the following property.
	
	\begin{proposition} 
		We have $\lim_{\gamma\to \infty} \text{ORCT}\left(\gamma\right) = \text{ODCT}$.
	\end{proposition}
	
	 In our numerical section, we will illustrate that with the logistic CDF family, and for large values of $\gamma$, ORCT yields better results than the ODCT reported in \cite{bertsimas2017optimal}. This means that, by just allowing a small level of randomization, corresponding to an almost deterministic cut, ORCT is preferable.
	
	We now prove limit results for ORCT when the sample $I=\left\lbrace\left(\bm{x}_i,y_i\right)\right\rbrace_{ 1\leq i \leq N}$ is independent and identically distributed (i.i.d.) and its size grows to infinity. 
	Unlike other optimization-based tree classifiers, the feasible region in \eqref{NLCO} does not depend on the training sample $I$, and the objective function is continuous and separable on $I$. Thus, Problem \eqref{NLCO} can be reformulated as the Sample Average Approximation (SAA) problem  of some theoretical or true stochastic problem, which makes it possible to show consistency of the estimators of the optimal value and the set of optimal solutions to their true counterparts, as stated in \cite{shapiro2009lectures}.
	
	We will start by rewriting \eqref{NLCO} into a more compact formulation. The decision variables $a_{jt},\,\,j=1,\ldots,p,\,\, t\in\tau_B$, $\mu_t,\,\, t\in\tau_B$ and $C_{kt},\,\, k=1,\ldots,K,\,\, t\in\tau_L$, are grouped into the $n$-dimensional decision vector $\bm{z}=\left(\bm{a},\bm{\mu},\bm{C}\right)^T$, where $n=\left( p+1\right) \vert\tau_B\vert + K\vert\tau_L\vert$. Note that $\bm{z}$ comprises information on the cuts in the branch nodes of the tree as well as the assignments in the leaf nodes. The feasible region will be denoted by $Z$, where \begin{align*}
	Z = \left\lbrace \bm{z}=\left(\bm{a},\bm{\mu},\bm{C}\right)^T\in\mathbb{R}^n: \right. &  \bm{a}\in\left[-1,1\right]^{p\times\lvert\tau_B\rvert}, \bm{\mu}\in\left[-1,1\right]^{\lvert\tau_B\rvert}, \sum_{k=1}^K C_{kt} = 1, t\in \tau_L,\\
	& \left. \sum_{t\in \tau_L} C_{kt} \geq 1, k=1,\ldots,K, {C_{kt} \geq 0,\,\,k=1,\ldots,K,\,\, t\in\tau_L}\right\rbrace
	\end{align*} which is a non-empty compact subset of $\mathbb{R}^n$. Then, we have a sample of $N$ i.i.d. realizations of a random vector $\bm{\xi}=\left( \bm{X},Y\right)$ whose probability distribution $E$ is supported on a set $\Xi\subset\mathbb{R}^{p+1}$,  i.e., we have $I=\left\lbrace \bm{\xi}_i= \left(\bm{x}_i,y_i\right)  \right\rbrace_{1\leq i\leq N}$. Thus, Problem \eqref{NLCO} can be written as:
	\begin{equation}
	\label{saa}
	\min_{\bm{z}\in Z} \,\, \left\lbrace \hat{g}_N\left(\bm{z}\right) := \dfrac{1}{N}\sum_{i=1}^N G\left(\bm{z},\bm{\xi}_i\right)\right\rbrace,
	\end{equation}  where \begin{equation}
	G\left(\bm{z},\bm{\xi}_i\right)= \sum\limits_{t\in\tau_L} P_{it}\left(\bm{a},\bm{\mu}\right) \sum\limits_{k=1}^K W_{y_ik}C_{kt}\label{g}.
	\end{equation}
	Problem \eqref{saa} can be seen as the SAA problem associated with the true stochastic problem:
	\begin{equation}
	\label{true}
	\min_{\bm{z}\in Z} \,\, \left\lbrace g\left(\bm{z}\right) := \mathbb{E} \left[G\left(\bm{z},\bm{\xi}\right)\right]\right\rbrace.
	\end{equation} That is, for any $\bm{z}\in Z$, the estimator of the expected value $g\left(\bm{z}\right)$, $\hat{g}_N\left(\bm{z}\right)$, is obtained by averaging values $G\left(\bm{z},\bm{\xi}_i\right),\,\,i=1,\ldots,N$.
	
	Let $\vartheta^*$ and $S$ denote the optimal value and the set of optimal solutions of the true problem \eqref{true}, respectively. Similarly, $\hat{\vartheta}_N$ and $\hat{S}_N$ will denote the optimal value and the set of optimal solutions of the SAA problem \eqref{saa}, respectively.  Our goal is to prove the consistency of the SAA estimators to the true counterparts. The estimator $\hat{\vartheta}_N$ of the parameter $\vartheta$ is said to be consistent, in the sense of \cite{shapiro2009lectures}, if $\hat{\vartheta}_N$ converges with probability $1$ (w.p.1) to $\vartheta$ as $N\to\infty$. For the set of optimal solutions, \cite{shapiro2009lectures} establish consistency of the estimator $\hat{S}_N$ to $S$ when the deviation of $\hat{S}_N$ from $S$, $\mathbb{D} (\hat{S}_N,S)$, converges w.p.1 to $0$ as $N\to\infty$, where $\mathbb{D} (\hat{S}_N,S)$ actually represents the distance between both sets.
	
	\begin{theorem}
		\label{consistency}
		$\hat{\vartheta}_N $ and $\hat{S}_N$ are consistent estimators of $\vartheta^*$ and $S$.
	\end{theorem}
	
	\noindent\textit{Proof of Theorem \ref{consistency}}.
	The result is a direct consequence of Theorems 7.48 and 5.3 in \cite{shapiro2009lectures}. We will first show that the conditions of Theorem 7.48 hold. Indeed, we have that the sample is i.i.d. and that $Z$ is a nonempty compact subset of $\mathbb{R}^n$. We also have that for any $\bm{z}\in Z$ the function $G\left(\cdot,\bm{ \xi}\right)$ is continuous at $\bm{z}$ for almost every $\bm{\xi}\in\Xi$, since the CDF $F$ is a continuous function by assumption. Finally, $G \left(\bm{z},\bm{\xi}\right),\,\, \bm{z}\in Z$, is dominated by the following constant function \[\left\lvert G\left(\bm{z},\bm{\xi}\right) \right\rvert\leq \max\limits_{k\neq y_i}\left\lbrace W_{y_i k}\right\rbrace.\] 
	This constant function is integrable since the support of the probability distribution $E$ is contained in a compact set, namely $\Xi\subset\left[0,1\right]^p \times\left[1,K\right]$. From Theorem 7.48, we know that $g\left(\bm{z}\right)$ is finite valued and continuous on $Z$ and $\hat{g}_N\left(\bm{z}\right)$ converges to $g\left(\bm{z}\right)$ w.p.1, as $N\to\infty$, uniformly in $\bm{z}\in Z$. In addition, we have that $\Hat{S}_N, S\subset Z$. This, together with the continuity of both $\hat{g}_N$ and $g$,  implies that $\Hat{S}_N, S\neq\emptyset$. Thus, the conditions of Theorem 5.3 hold, and the desired result follows.\hfill $\square$\\

	\section{Variants of ORCTs}\label{Variants of ORCTs}
	
	In this section, two variants of ORCT are discussed. The first one is an extension to ORCT that allows one to control the expected classification performance in a given class; while the second one is the randomized optimal version of a regression tree.
	
	\subsection{ORCT with constraints on expected performance}
	\label{ORCT with constraints on expected performance}
	
		Although classifiers seek  a rule yielding a good overall classification rate, there are many cases in which misclassification has different consequences for different classes. It is then
		more appealing not to focus on the overall classification, but to obtain an acceptable overall performance while ensuring a certain level of performance in some classes. Our ORCT has two ways of achieving this. First, one could modify the misclassification costs \eqref{w} in the objective function \eqref{objective2},  for different sets of values of $W_{y_ik}$. However, in this way, one has no direct control on the misclassification rates for critical classes. Second, ORCT is flexible enough to allow the incorporation of constraints on expected performance over the different classes explicitly. Indeed, define the random variable $O_i$,
	\begin{equation}
	\nonumber
	O_i= \left\{ \begin{array}{ll}
	1, & \text{if individual $i$ is correctly classified} \\
	0, & \text{otherwise}.
	\end{array}
	\right.
	\end{equation}
	
	Given $k=1,\ldots,K$, a Correct Classification Rate (CCR) over the $k-$th class, namely, $\rho_k$, is desired:
	\begin{equation}
	\nonumber
	\dfrac{1}{|I_k|}\sum_{i\in I_k} O_i \geq \rho_k.
	\end{equation}

	The expectation of achieving this performance can be written as:
	\begin{equation}
	\nonumber
	\mathbb{E}\left[ \dfrac{1}{|I_k|}\sum_{i\in I_k} O_i \right] =\dfrac{1}{|I_k|} \sum_{i \in I_k} \mathbb{E}\left[ O_i\right] = \dfrac{1}{|I_k|}\sum_{i\in I_k} \sum_{t\in\tau_L} P_{it}\left(\bm{a},\bm{\mu}\right)C_{kt}\geq \rho_k.
	\end{equation} 
	Hence, given the class $k=1,\ldots,K$ to be controlled and its desired performance $\rho_k$, the following constraint would need to be added to the model:
	\begin{equation}
	\label{performanceconstraints}
	\sum_{i\in I_k} \sum_{t\in\tau_L} P_{it}\left(\bm{a},\bm{\mu}\right)C_{kt} \geq \rho_k |I_k|.
	\end{equation}
	
	\subsection{Optimal Randomized Regression Trees}
	\label{Optimal Randomized Regression Trees}
	
	Up to now, we have focused on fitting classification trees. Yet, the same randomized framework can be considered for the regression task, where the outcome is no longer discrete but continuous-valued, i.e., $y_i\in\mathbb{R}$.
	
	An Optimal Randomized Regression Tree (ORRT) constructs oblique cuts on every branch node and evaluates them using a CDF. With this, we have the probability of each individual in the sample falling into every leaf node, as in Equation \eqref{c2}. The main difference lies in the leaf nodes, where the decision variables $\bm{C}=(C_{k t})_{k=1,\ldots,K,t\in \tau_L}$, are replaced by new decision variables $\bm{\varphi}=\left(\varphi_t\right)_{t\in\tau_L}$, that model the outcome value associated with leaf node $t$.
	
	For fixed $\bm{a},\bm{\mu},\bm{\varphi}$, the estimated outcome value for individual $i$ would be the average, over the leaf nodes, of the outcome values, $\varphi_t,\,\,t\in\tau_L$, weighted by the probability of belonging to such leaf node: 
	\begin{equation}
	\nonumber
	\sum_{t\in\tau_L}  P_{it}\left(\bm{a},\bm{\mu}\right) \varphi_t,
	\end{equation} where $P_{it}\left(\bm{a},\bm{\mu}\right)$ is defined as in \eqref{c2}.
	
	As customary in regression, the goal of ORRT is to minimize the mean squared error. As a consequence, ORRT can also be formulated as an NLCO problem as follows: 
	\begin{align}
	\nonumber
	\begin{split}
{\min}\hspace*{0.6cm} & \dfrac{1}{N} \sum_{i=1}^N\left(\sum_{t\in\tau_L} P_{it}\left(\bm{a},\bm{\mu}\right) \varphi_t-y_i\right)^2\\
	\text{s.t. \hspace*{0.5cm}} & a_{jt} \in \left[-1,1\right],\,\, j=1,\ldots,p,\,\, t\in \tau_B,\\
	& \mu_{t} \in \left[-1,1\right],\,\, t\in \tau_B,\\
	& \varphi_{t} \in \mathbb{R}, \,\, t\in \tau_L.
	\end{split}\label{ORRT}
	\end{align} 

		\section{Computational experiments}
		\label{Computational experiments}
		
		The purpose of this section is to illustrate the performance of our ORCT, against the natural benchmarking tree-based methods.
		
		We can draw the following conclusions from our computational experiments. First, in terms of classification accuracy, our approach outperforms CART{, OC1} and a benchmark decision tree approach based on integer programming, {is comparable to \textit{oblique.tree},} and is close to the local-search heuristic approach in \cite{dunn2018optimal} and Random Forests. Second, we show that our running times are low. Third, we illustrate the flexibility of ORCTs to produce variable importance metrics, and to handle cost-sensitive constraints, {unlike heuristic approaches.}
		
		Several well-known data sets from the UCI Machine Learning Repository \citep{Lichman:2013} have been chosen for the computational experiments. Table \ref{Information about the data sets considered.} reports the size, the number of predictor variables, and the number of classes as well as the corresponding class distribution. This selection comprises data sets of different nature: from small-sized data sets, such as Iris, to larger data sets, such as Thyroid-disease-ann-thyroid, Spambase and Magic-gamma-telescope; the class distribution is also diverse, from Iris or Seeds, which are balanced data sets, to Ozone-level-detection-one, which is highly imbalanced; lastly, and in addition to numerical (continuous as well as integer) predictor variables, we have also considered data sets with categorical predictor variables, modeled, as usual, through dummies.
		
		\begin{table}
			\centering
			{
			\caption{Information about the data sets considered.}
			\label{Information about the data sets considered.}
			\begin{tabular}{ l  c  c  c  c  c  c }
				\hline			
				Data set & Abbreviation & $N$ & $p$ & $K$ & Class distribution\\ \hline
				Connectionist-bench-sonar & CBS & 208 & 60 & 2 & 55\% - 45\% \\
				\hline
				Wisconsin & W & 569 & 30 & 2 &63\% - 37\%\\
				\hline
				Credit-approval & CA & 653 & 37 & 2 & 55\% - 45\% \\
				\hline
				Pima-indians-diabetes & PID & 768 & 8& 2& 65\% - 35\% \\
				\hline
				Statlog-project-German-credit & SPGC & 1000 & 48 & 2 & 70\% - 30\% \\
				\hline
				Ozone-level-detection-one & OLDO & 1848 & 72 & 2 & 97\% - 3\% \\
				\hline
				Spambase & SB & 4601 & 57 & 2& 61\% - 39\% \\
				\hline
				Magic-gamma-telescope & MGT & 19020 & 10 & 2 & 65\% - 35\% \\ \hline
				Iris & I & 150 & 4 & 3 & 33.3\%-33.3\%-33.3\% \\
				\hline
				Wine & Wi & 178 & 13 & 3 & 40\%-33\%-27\% \\
				\hline
				Seeds & S & 210 & 7 & 3 & 33.3\%-33.3\%-33.3\% \\
				\hline
				Thyroid-disease-ann-thyroid & TADT & 3772 & 21 & 3 & 92.5\%-5\%-2.5\% \\
				\hline
				Car-evaluation & CE & 1728 & 15 & 4 & 70\%-22\%-4\%-4\% \\
				\hline
			\end{tabular}
		}
		\end{table}
		
		The NLCO model \eqref{NLCO} has been implemented using Pyomo optimization modeling language \citep{hart2011pyomo,hart2017pyomo} in Python 3.7 \citep{pthn}. As a solver, we have used IPOPT 3.11.1 \citep{Wachter2006}. 
		 Our experiments have been conducted on a PC, with an Intel$^\circledR$ Core$^{\rm TM}$ i7-7700 CPU 3.60GHz processor and 32 GB RAM. The operating system is 64 bits.
		
		To train the ORCT, we solve the NCLP problem $20$ times, starting from different random initial solutions.
		
		The CDF chosen has been the logistic one, see Equation \eqref{logisticCDF}. In our computational experience, we illustrate that a small level of randomization is enough for obtaining good results. Thus, we have set $\gamma=512$ for both constructing and testing our model.
		
		Equal misclassification weights, $W_{y_i k} = 0.5,\,\, i=1,\ldots,N,\,\, k=1,\ldots,K,\,\, k\neq y_i$, have been used for the experiments.
		
		Each data set has been split into two subsets: the training subset (75\%) and the test subset (25\%). The ORCT is built over the training subset and, then, its performance is evaluated by determining the out-of-sample accuracy over the test subset. This procedure has been repeated ten times, and average results are reported.
		
		We will compare our ORCT of depths $D=1,\ldots,4$ with: the full CART, as implemented in the \texttt{rpart} R package \citep{rpartR}; {two other greedy approaches that implement oblique cuts: the full OC1 and the full \textit{oblique.tree}, as implemented in \cite{murthy1994system} and \cite{gitOT}, respectively, with the default tuning parameters;} the OCT-H MIO, proposed in \citep{bertsimas2017optimal} that also constructs oblique cuts, at the same depth as ORCT; the OCT-H LS in \cite{dunn2018optimal} that employs a local-search heuristic for building oblique trees at maximum depth $D=10$; and Random Forests from \texttt{randomForest} R package \citep{rfR} with the default tuning parameters.
		
		\subsection{Results for ORCT}
		\label{Results for ORCT}
		
		Tables \ref{$D=1$.}, \ref{$D=2$.}, \ref{$D=3$.} and \ref{$D=4$.} present the comparison of ORCT at depth $D=1,2,3$ and $4$, respectively, against the benchmarks described above. {Figures \ref{plotD1f}, \ref{plotD2f}, \ref{plotD3f} and \ref{plotD4f} depict these results, where ORCT is highlighted in striped grey.} Note that for data sets with $K\geq 3$, the ORCT at depth $D=1$ would become infeasible due to the set of constraints \eqref{strenght}.
		
	{Each table and each figure displays, per data set, the average out-of-sample accuracy over the ten runs for CART{, OC1, \textit{oblique.tree}}, RF and ORCT, as well as the average out-of-sample accuracy across all data sets. Results for OCT-H MIO and OCT-H LS are taken from \citep{bertsimas2017optimal} and \citep{dunn2018optimal}, respectively. Note that the Magic-gamma-telescope data set is not available in \citep{bertsimas2017optimal}. Tables also include information about the average execution time for ORCT. This time involves the data reading, the scaling of the training set, the random generation of initial solutions, the optimization time, the scaling of the test set using the scale parameters obtained in the training and the evaluation of performance. The average solving time of an instance of a specific optimization problem is also shown. Moreover, for each data set, we rank the methods by their accuracy. The rank is shown in parentheses. A rank of $1$ indicates that the method is the best in terms of out-of-sample accuracy on a given data set and a rank of {7} indicates that the method performed the worst. The average accuracy and rank of each method across all data sets are found at the bottom of the table.		}

\begin{table}
	\caption{Results for $D = 1$ in terms of the out-of-sample accuracy.}
	\label{$D=1$.}
	\centering 
	\hspace*{-0.75cm}
	{\small 
		\begin{tabular}{ l c c c c c c c c}\hline
			\multirow{2}{*}{Data set}   & \multicolumn{7}{c}{Out-of-sample accuracy} & Average time \\\cline{2-8}
		 & CART & {OC1} & {\textit{oblique.tree}}  & OCT-H MIO & OCT-H LS & RF  & ORCT 	& (in secs) \\\hline
			CBS & 70.0(7) & { 70.8(5)} & { 72.5(4)}  & 70.4(6) & 77.3(2) & 83.1(1)& 76.3(3)  & 8 \\ \hline
			W   & 92.0(7) & {94.1(4)} & {93.7(5)}  & 93.1(6) & 94.8(3) & 95.5(2)& 96.4(1) & 10\\ \hline
			CA  & 85.7(4) &{82.0(7)} & {83.5(6)}   & 87.9(1) & 86.0(3) & 86.7(2) & 83.7(5) & 7\\ \hline
			PID  & 74.2(4)&{60.7(7)} & {76.0(2)}   & 71.6(6) & 73.1(5) & 76.3(1) & 76.0(2)& 6\\ \hline
			SPGC & 72.1(3) &{68.5(7)} & {73.8(2)}   & 71.6(6) & 72.1(3) & 75.2(1) & 72.1(3)  & 10\\ \hline
			OLDO  & 95.6(5) & {95.5(6)} & {92.9(7)}  & 96.8(1) & 96.2(4) & 96.4(3)& 96.7(2) & 66\\ \hline
			SB  & 89.2(6) & {92.3(4)} & {92.7(3)}   & 83.6(7) & 94.2(2) & 95.1(1) & 89.8(5)& 49\\ \hline
			MGT & 82.1(4) &{ 78.8(6)} & {82.7(3)}  & - & 86.9(2) & 87.7(1)& 79.9(5) & 122\\ \hline \hline
			Average &82.6(5.0) & {80.3(5.8)}  & {83.5(4.0)} & 82.1(4.7)& 85.1(3.0)& 87.0(1.5)& 83.9(3.3)& 35 \\		
	\end{tabular}}
\end{table} 

\begin{figure}
\hspace{-1cm}
\includegraphics[scale=0.45]{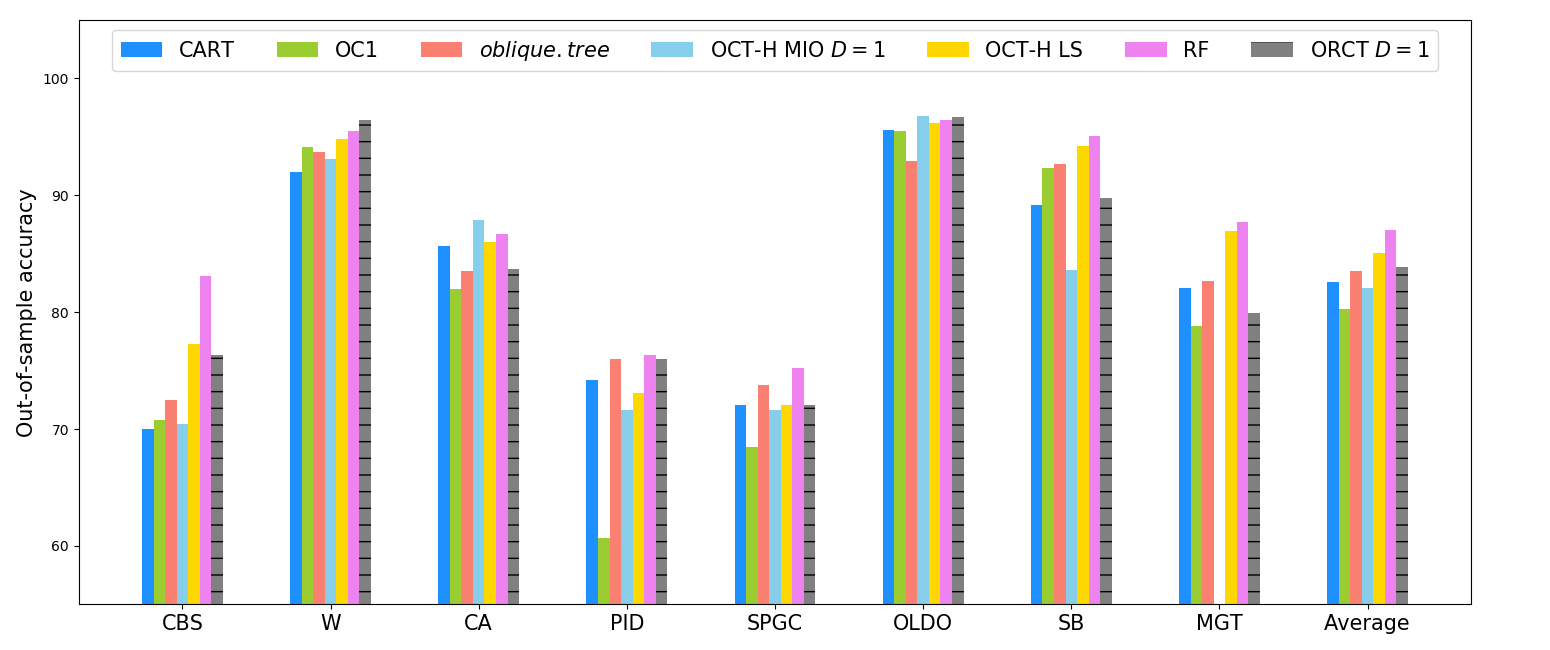}
\caption{Comparison of ORCT at depth $D=1$ and other tree-based methods in terms of the out-of-sample accuracy.}\label{plotD1f}
\end{figure}
	
		We start by discussing the results for $D=1$ in Table \ref{$D=1$.} {and \figurename{ \ref{plotD1f}}}. {We will say that two methods have a comparable accuracy if they differ in less than $1$ percentage point.} ORCT generally outperforms CART, even with depth $D=1$, i.e., one single oblique cut. This is the case for all data sets except for CA and MGT. {Regarding to OC1, ORCT is generally better except for SB. With respect to \textit{oblique.tree}, ORCT is comparable (CA, PID) or better (CBS, W, OLDO) in five out of eight datasets.  ORCT is generally better than OCT-H MIO except for CA.} With respect to OCT-H LS, ORCT outperforms {or comparable} in four out of eight data sets (W, PID, SPGC, OLDO). Finally, RF tends to lead in performance among all one-single-tree-based methods we have tested. Nonetheless, ORCTs are comparable to RFs in some data sets (PID, OLDO) and even competitive in others (W). In terms of the average performance across all data sets, ORCT is superior to CART{, OC1} and OCT-H MIO{, and is comparable to \textit{oblique.tree}}. Although OCT-H LS has a higher average accuracy across all data sets than ORCT, they both have the same average rank. RF presents the best average accuracy as well as rank.

				\begin{table}
			\caption{Results for $D = 2$ in terms of the out-of-sample accuracy.}
			\label{$D=2$.}
			\centering 
			\hspace*{-0.75cm}
			{\small
			\begin{tabular}{ l c c c c c c c c}\hline
				\multirow{2}{*}{Data set}   & \multicolumn{7}{c}{Out-of-sample accuracy}& Average time \\\cline{2-8}
			 & CART & {OC1}& {\textit{oblique.tree}}  & OCT-H MIO & OCT-H LS & RF  & ORCT	& (in secs)\\\hline
				CBS   & 70.0(6)& {70.8(5)}& {72.5(4)}& 70.0(6) & 77.3(3) & 83.1(1)  & 77.5(2) & 27\\ \hline
				W   & 92.0(7)& {94.1(4)}& {93.7(5)} & 93.1(6) & 94.8(3) & 95.5(2)  & 96.2(1) & 36 \\ \hline
				CA  & 85.7(4)& {82.0(7)}& {83.5(6)}  & 87.9(1) & 86.0(3) & 86.7(2) & 84.2(5)& 20\\ \hline
				PID  & 74.2(4)& {60.7(7)}& {76.0(2)} & 71.4(6) & 73.1(5) & 76.3(1) &  76.0(2)& 20\\ \hline
				SPGC  & 72.1(4)& {68.5(7)} & {73.8(2)} & 70.4(6)  & 72.1(4) & 75.2(1)&  72.8(3) & 40\\ \hline
				OLDO  & 95.6(5)& {95.5(6)}& {92.9(7)}  &  96.8(1) & 96.2(4) & 96.4(3) &  96.7(2)& 267\\ \hline
				SB  & 89.2(6)& {92.3(4)}& {92.7(3)} & 85.7(7) & 94.2(2) & 95.1(1)  & 89.8(5)& 58\\ \hline
				MGT  & 82.1(4)& {78.8(6)}& {82.7(3)}  & - & 86.9(2) & 87.7(1)& 80.8(5)& 551\\ \hline
				I  & 92.7(7)& {95.4(3)}& {96.8(1)}  & 95.1(5) & 94.6(6) & 95.4(3)& 95.9(2)& 5  \\ \hline
				Wi  & 88.6(7)& {90.9(6)}& {96.1(3)}   & 91.1(5) & 95.1(4) & 98.6(1) & 96.6(2)& 9\\ \hline
				S  & 90.2(7)& {90.4(6)}& {94.2(1)} & 90.6(5) & 91.7(4) & 92.5(3)  & 94.2(1)& 7\\ \hline
				TDAT & 99.1(2)& {97.9(4)} & {97.7(5)}  & 92.5(6) & 99.7(1) & 99.1(2) & 92.2(7) & 111\\ \hline
				CE & 88.1(5)& {94.7(2)}& {93.9(3)}  & 87.5(7) & 97.8(1) & 88.0(6) & 89.8(4) & 42\\ \hline \hline
				Average  & 86.1(5.2)& {85.5(5.1)}& {88.1(3.5)}& 86.0(5.1) & 89.2(3.2)&90.0(2.1)& 87.9(3.2)& 92 \\
			\end{tabular}}
		\end{table}

\begin{figure}
\hspace{-1cm}
\includegraphics[scale=0.45]{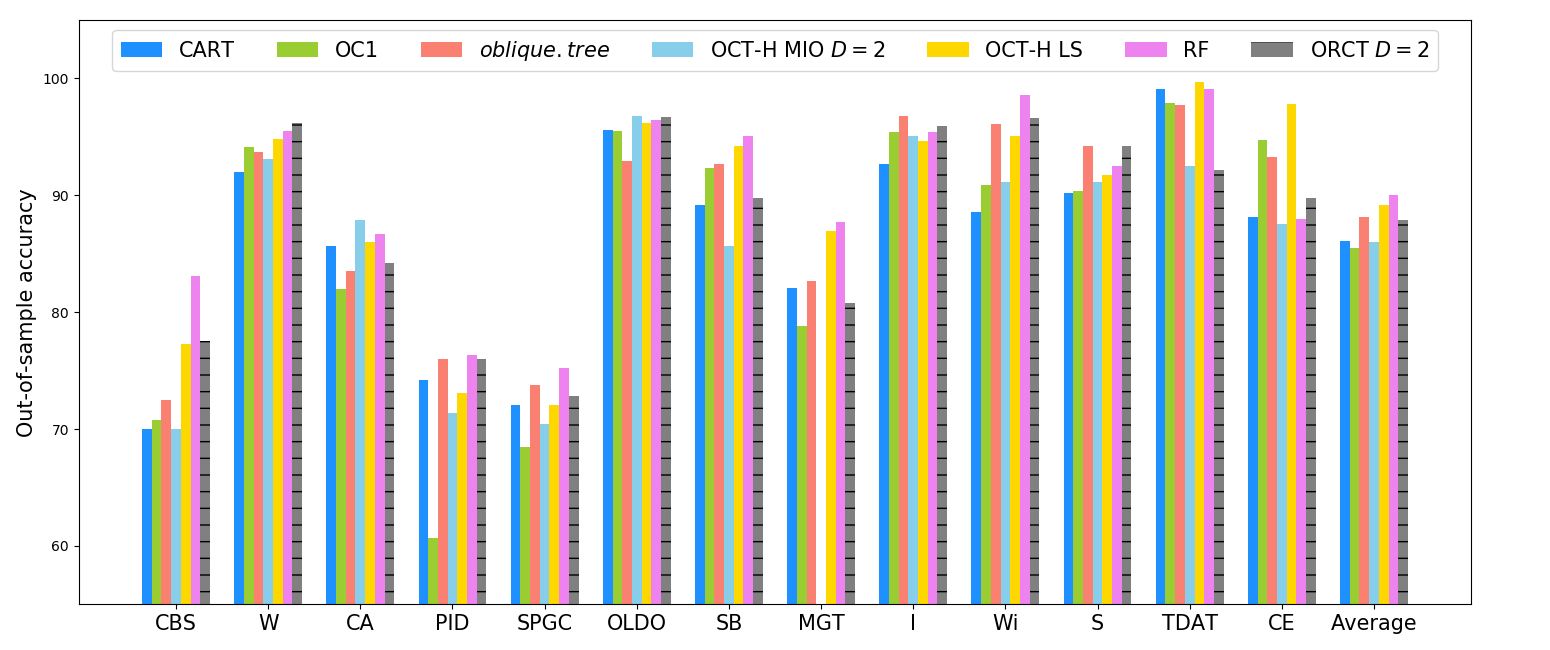}

\caption{Comparison of ORCT at depth $D=2$ and other tree-based methods in terms of the out-of-sample accuracy.} \label{plotD2f}
\end{figure}
	
		We continue by discussing the results for $D=2$ in Table \ref{$D=2$.} { and \figurename{ \ref{plotD2f}}}. The out-of-sample accuracy has improved for both CA and MGT, in comparison with $D=1$, but the same conclusions as in Table \ref{$D=1$.} can be drawn for two-class problems when comparing ORCT with CART and OCT-H MIO. The outperformance of ORCT against CART and OCT-H MIO is also clear for I, W, S and CE. This is not the case for TDAT, in which we are comparable to OCT-H MIO although CART outperforms these two methods. { Compared to OC1, ORCT is comparable (I) or outperforms (the rest except for SB, TDAT and CE) in ten out of the thirteen datasets. With respect to \textit{oblique.tree}, ORCT is comparable (CA, PID, I, W, S) or outperforms (CBS, W, OLDO) in eight out of the thirteen data sets.} Regarding to OCT-H LS, ORCT is competitive (CBS) or outperforms (W, PID, SPGC, OLDO, I, W, S) in eight out of thirteen data sets. Finally, we are again comparable to RF in some of the data sets (PID, OLDO) and even competitive in others (W, I, Wi, S). On average across all data sets, RF is the best method and ORCT is superior to CART and OCT-H MIO again{, and is comparable to \textit{oblique.tree}}. A slightly greater average rank is observed for ORCT at depth $D=2$, although OCT-H LS still produces a higher average accuracy.

\begin{table}
	\caption{Results for $D = 3$ in terms of the out-of-sample accuracy.}
	\label{$D=3$.}
	\centering 
	\hspace*{-0.75cm}
	{\small
	\begin{tabular}{ l c c c c c c c c}\hline
		\multirow{2}{*}{Data set} & \multicolumn{7}{c}{Out-of-sample accuracy}  & Average time  \\\cline{2-8}
	 & CART & {OC1} & {\textit{oblique.tree}}   & OCT-H MIO & OCT-H LS & RF & ORCT 	& (in secs)\\\hline
		CBS & 70.0(7)& {70.8(5)} & {72.5(4)} & 70.8(5) & 77.3(2) & 83.1(1)& 77.1(3) &  84\\ \hline
		W  & 92.0(7)&{94.1(4)}  & {93.7(6)}  & 94.0(5) & 94.8(3) & 95.5(2) & 96.4(1)& 107\\ \hline
		CA   & 85.7(4)& {82.0(7)} & {83.5(6)} & 87.9(1) & 86.0(3) & 86.7(2) & 84.4(5)& 62\\ \hline
		PID & 74.2(4)&{60.7(7)} & {76.0(2)} & 71.4(6) & 73.1(5) & 76.3(1) &  75.2(3) & 73\\ \hline
		SPGC  & 72.1(4)&{68.5(7)}  & {73.8(2)} & 71.0(5)  & 72.1(4) & 75.2(1) &  73.4(3)& 116\\ \hline
		OLDO & 95.6(5)& {95.5(6)} & {92.9(7)}  &  96.8(1) & 96.2(4) & 96.4(3) &  96.7(2) & 918\\ \hline
		SB  & 89.2(6)& {92.3(4)} & {92.7(3)}  & 86.6(7) & 94.2(2) & 95.1(1)& 89.8(5)& 572 \\ \hline
		MGT & 82.1(5)& {78.8(6)} & {82.7(4)}  & - & 86.9(2) & 87.7(1)& 82.9(3) & 2018\\ \hline
		I  & 92.7(7) & {95.4(4)}  & {96.8(1)}& 95.1(5) & 94.6(6) & 95.4(3) & 95.7(2)&  12\\ \hline
		Wi & 88.6(7)& {90.9(6)}  & {96.1(3)}  & 92.9(5) & 95.1(4) & 98.6(1) & 96.6(2) & 26\\ \hline
		S & 90.2(7)& {90.4(6)} & {94.2(1)} & 91.3(5) & 91.7(4) & 92.5(3)  & 94.0(2) & 22\\ \hline
		TDAT  & 99.1(2)& {97.9(4)} & {97.7(5)}   & 92.5(6) & 99.7(1) & 99.1(2) & 92.5(7) & 367\\ \hline
		CE  & 88.1(5)& {94.7(2)} & {93.9(3)}  & 87.5(7) & 97.8(1) & 88.0(6)& 91.7(4)& 161 \\ \hline \hline
		Average  & 86.1(5.4)& {85.5(5.2)} & {88.1(3.6)} & 86.5(4.8) & 89.2(3.1) & 90.0(2.1) & 88.2(3.2)& 349\\
	\end{tabular}}
\end{table}

\begin{figure}
\hspace{-1cm}
\includegraphics[scale=0.45]{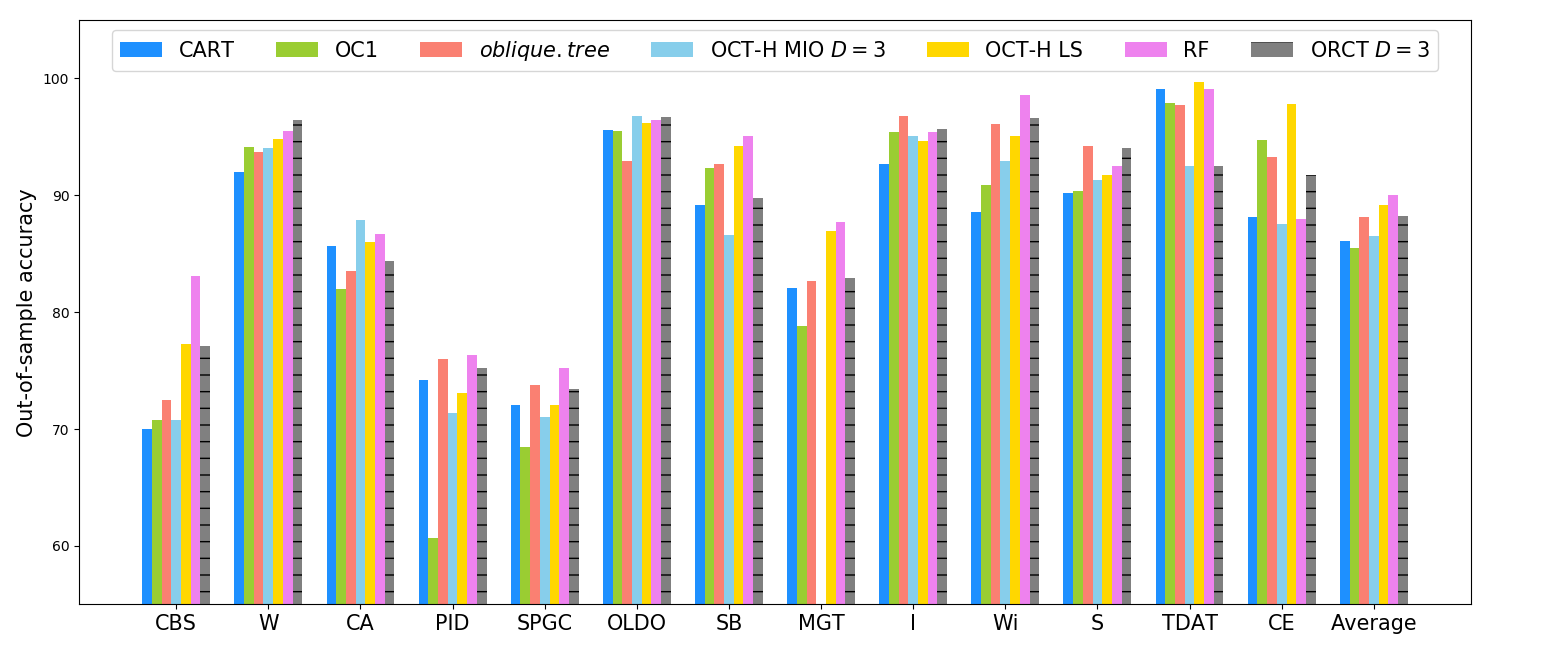}
\caption{Comparison of ORCT at depth $D=3$ and other tree-based methods in terms of the out-of-sample accuracy.}
\label{plotD3f}
\end{figure}

		Similar conclusions can be drawn for $D=3$ and $D=4$, see Tables \ref{$D=3$.} and \ref{$D=4$.}, { and Figures \ref{plotD3f} and \ref{plotD4f}}, respectively. The out-of-sample accuracies of ORCT at depths $D=3$ and $4$ have not significantly changed in most cases except for CE and MGT. { The out-of-sample accuracies for both are already comparable to \textit{oblique.tree} and superior to CART, respectively.} The comparisons between { ORCT and OC1,} ORCT and OCT-H LS, and ORCT and RF remain the same. On average across all data sets, ORCT, while maintaing a slightly better average rank, tends to close its gap with OCT-H LS regarding the average out-of-sample accuracy across all data sets. 
		
\begin{table}
	\caption{Results for $D=4$ in terms of the out-of-sample accuracy.}
	\label{$D=4$.}
	\centering 
	\hspace*{-0.75cm}
	{\small
	\begin{tabular}{ l c c c c c c c c}\hline
		\multirow{2}{*}{Data set}   & \multicolumn{7}{c}{Out-of-sample accuracy} & Average time\\\cline{2-8}
	 & CART & {OC1} & {\textit{oblique.tree}}   & OCT-H MIO & OCT-H LS & RF& ORCT	& (in secs) \\\hline
		CBS  & 70.0(7)& {70.8(6)}& {72.5(4)}  & 71.5(5) & 77.3(2) & 83.1(1) & 76.5(3)& 210\\ \hline
		W  & 92.0(7)& {94.1(4)}& {93.7(6)}   & 94.0(5) & 94.8(3) & 95.5(2) & 96.2(1)& 351\\ \hline
		CA  & 85.7(4)& {82.0(7)}& {83.5(6)} & 87.9(1) & 86.0(3) & 86.7(2) & 84.6(5) & 203\\ \hline
		PID  & 74.2(4)& {60.7(7)}& {76.0(3)}& 70.3(6) & 73.1(5) & 76.3(1) &  76.1(2) & 208\\ \hline
		SPGC  & 72.1(4)& {68.5(7)}& {73.8(2)} & 71.0(6)  & 72.1(4) & 75.2(1)&  72.8(3) & 415\\ \hline
		OLDO  & 95.6(5)& {95.5(6)}& {92.9(7)} &  96.8(1) & 96.2(4) & 96.4(3)  &  96.7(2)& 3360\\ \hline
		SB  & 89.2(6)&{92.3(4)}& {92.7(3)}  & 86.6(7) & 94.2(2) & 95.1(1) & 89.8(5)& 1717\\ \hline
		MGT  & 82.1(5) & {78.8(6)}& {82.7(4)} & - & 86.9(2) & 87.7(1)& 84.3(3)& 5603\\ \hline
		I & 92.7(7)& {95.4(3)} & {96.8(1)} & 95.1(5) & 94.6(6) & 95.4(3) & 96.2(2)&  31 \\ \hline
		Wi & 88.6(7)& {90.9(6)} & {96.1(2)} & 91.6(5) & 95.1(4) & 98.6(1)& 95.7(3)   & 69\\ \hline
		S  & 90.2(7)& {90.4(6)}& {94.2(1)} & 91.3(5) & 91.7(4) & 92.5(3) & 93.1(2) & 58\\ \hline
		TDAT  & 99.1(2)& { 97.9(4)}& {97.7(5)}  & 92.5(7) & 99.7(1) & 99.1(2) & 93.1(6)& 1051\\ \hline
		CE  & 88.1(5)&{94.7(2)}& {93.3(4)} & 87.5(7) & 97.8(1) & 88.0(6)  & 93.6(3)& 468\\ \hline \hline
		Average  & 86.1(5.4) & { 85.5(5.2)}& {88.1(3.7)} & 86.3(5.0) & 89.2(3.1) & 90.0(2.1) & 88.4(3.1)& 1057\\
	\end{tabular}}
\end{table}

\begin{figure}
\hspace{-1cm}
\includegraphics[scale=0.45]{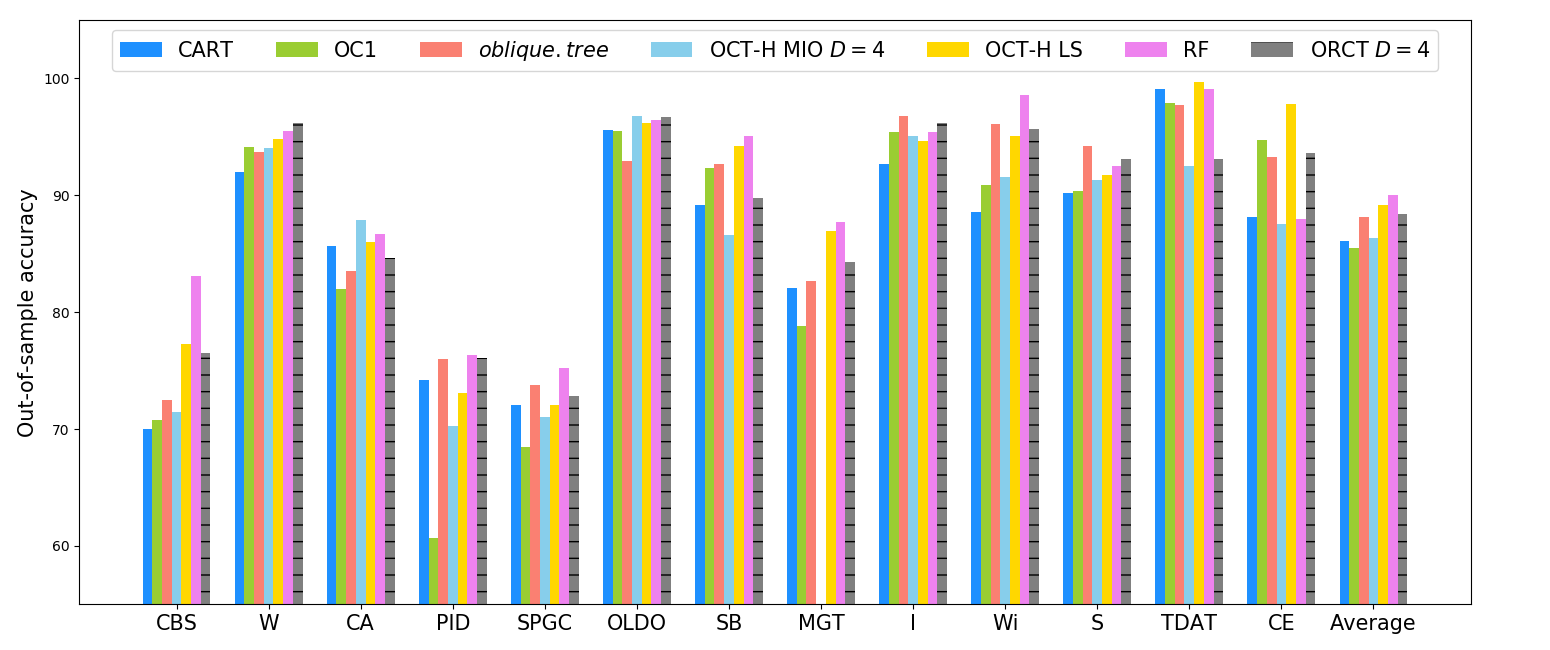}
\caption{Comparison of ORCT at depth $D=4$ and other tree-based methods in terms of the out-of-sample accuracy.}
\label{plotD4f}	
\end{figure}

		In summary, these numerical results illustrate that, in terms of accuracy, ORCT outperforms CART{, OC1} and OCT-H MIO, while ORCT { is comparable to \textit{oblique.tree} and} manages to get close to OCT-H LS and RFs. In contrast to these { three} heuristic approaches, ORCT has a direct and effective control on critical issues such as cost-sensitiveness, as will be seen in Section \ref{Results for ORCT with constraints on expected performance}.
		
		Regarding the computational time taken by ORCT, Tables \ref{$D=1$.}, \ref{$D=2$.}, \ref{$D=3$.} and \ref{$D=4$.} report low running times, compared to OCT-H MIO, where a CPU time limit from $5$ to $15$ minutes was imposed in \cite{bertsimas2017optimal}; excluding the computing time devoted to preprocessing that involves the tuning of a parameter for which time limits of 60 seconds were imposed to OCT-H MIO, which is not the case for us. OCT-H LS also requires parameters tuning. Although limited information about computing times is found in \cite{dunn2018optimal}, a large gain in time is reported for a particular data set, compared to OCT-H MIO.

	\subsection{Results for variable importance measures via ORCTs}
	\label{Results for variable importance measures via ORCTs}
	
	Measuring the importance of predictor variables in RFs has been thoroughly studied in the literature. In particular, the two most popular importance measures for forests are outlined in \cite{biau2016random}: the Mean Decrease Impurity (MDI) and the Mean Decrease Accuracy (MDA). The MDI takes advantage of the splitting criterion used for growing classic decision trees: given an impurity measure, the predictor variable that maximizes the decrease of impurity together with its corresponding splitting threshold are chosen. Thus, the MDI of predictor variable $j$ is the average over every tree built of the decrease in impurity of splits along that variable, weighted with the fraction of individuals falling in the corresponding branch node. The MDA is based on the following notion: if a predictor variable is not influential, permuting the values it takes should not affect the prediction accuracy of the forest. Thus, the MDA of predictor variable $j$ is the average over every tree built of the difference in accuracy before and after the permutation.
	
	In our case we have two straightforward and inexpensive ways of measuring the importance of predictor variables by analyzing the distribution of the absolute values of coefficients $\bm{ a_{jt}}$. The first metric to measure the importance of predictor variable $j,\,\,j=1,\ldots,p$, is to sum the absolute values of the coeffientes $a_{jt}$ for all branch nodes $t\in\tau_B$, called hereafter the Sum Importance Measure (SIM):
	\begin{equation}
	\nonumber
	\text{SIM}_j = \sum_{t\in \tau_B} \vert a_{jt} \vert.
	\end{equation}
	The second metric is the so-called Maximum Importance Measure (MIM), which takes instead the maximum among all the branch nodes $t\in\tau_B$:
	\begin{equation}
	\nonumber
	\text{MIM}_j = \max_{t\in \tau_B} \vert a_{jt} \vert.
	\end{equation}
	For illustrative purposes, these variable importance measures have been evaluated for the Wine and Car-evaluation data sets in Table \ref{Information about the data sets considered.}, see Figures \ref{impwine} and \ref{impcar}. Both measures have been evaluated over the resultant ORCT of the tenth run. The variable importance measures for RFs, the MDA and the MDI using the Gini index as the impurity measure, are also depicted for the aforementioned data sets. These can be found in Figures \ref{impwinerf} and \ref{impcarrf} and have been obtained with the \texttt{randomForest} R package. The message conveyed by these plots is, in general, in agreement. For instance, V8 in the Wine data set is not as important as V10. The same conclusion can be drawn for the Car-evaluation data set.
	
	\begin{figure}
		\centering
		\includegraphics[height=4.2cm,width=6.6cm]{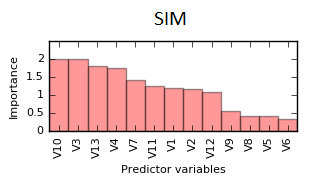}
		\includegraphics[height=4.4cm]{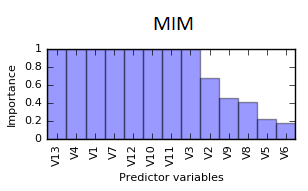}
		\caption{ORCT variable importance measures for the Wine data set.}
		\label{impwine}
	\end{figure}
	
	\begin{figure}
		\centering
		\includegraphics[scale=1]{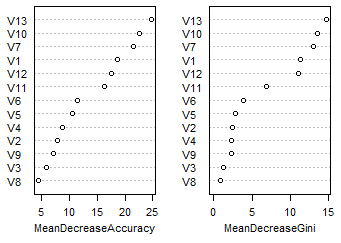}
		\caption{RF variable importance measures for the Wine data set.}
		\label{impwinerf}
	\end{figure}
	
	\begin{figure}
		\centering
		\includegraphics[height=4.5cm,width=6.6cm]{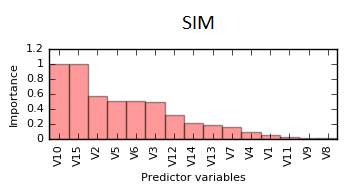}
		\includegraphics[height=4.5cm,width=6.6cm]{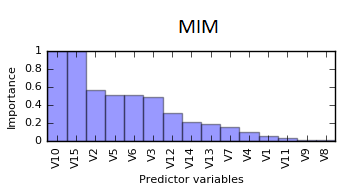}
		\caption{ORCT variable importance measures for the Car-evaluation data set.}
		\label{impcar}
	\end{figure}
	
	\begin{figure}
		\centering
		\includegraphics[scale=1]{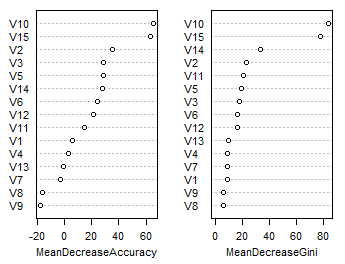}
		\caption{RF variable importance measures for the Car-evaluation data set.}
		\label{impcarrf}
	\end{figure}

	\subsection{Results for ORCT with constraints on expected performance}
	\label{Results for ORCT with constraints on expected performance}
	
	The Pima-indians-diabetes data set, from \cite{Lichman:2013}, consists of a sample of 768 patients of Pima Indian heritage in the USA, on which 8 predictor variables are measured. The target variable is whether the patient shows signs of diabetes or not. Diabetics are the positive class and represent the 35\% of the entire sample. Firstly, we have run our ORCT without constraints on expected performance by imposing a correct classification percentage over the positive class, $\rho_{+}$, equal to zero. See the first row in Table \ref{Results with constraints on expected performance over the positive class in the Pima-indians-diabetes data set.}. 
	
	\begin{table}[H]
		\caption{Results with constraints on expected performance over the positive class in the Pima-indians-diabetes data set.}
		\label{Results with constraints on expected performance over the positive class in the Pima-indians-diabetes data set.}
		\centering\hspace{1cm}
		\begin{tabular}{ c c c c c c c}
			\hline
			Imposed $\textit{TPR} \,\, (\rho_{+})$ & $\textit{TPR}_{\textit{train}}$ & $\textit{TPR}_{\textit{test}}$ & $\textit{TNR}_{\textit{train}}$ & $\textit{TNR}_{\textit{test}}$ & $\textit{CCR}_{\textit{train}}$ & $\textit{CCR}_{\textit{test}}$ \\
			\hline
			0  & 60.8 & 56.4 & 90.5 & 87.8 & 80.3 & 76.2 \\ \hline
			62.5&63.9&59.0& 88.6&86.0&80.1&76.1 \\ \hline
			65.0&65.8&60.9&87.3&84.7&79.9&75.9 \\ \hline
			67.5&68.4&62.5&85.5&83.1&79.6&75.5 \\ \hline
			70.0&71.1&64.3&83.6&81.8&79.3&75.3 \\ \hline
			72.5&73.7&67.4&81.4&80.1&78.7&75.4 \\ \hline
			75.0&75.8&68.1&79.3&77.4&78.1&73.9 \\ \hline
			77.5&78.9&72.9&77.2&75.9&77.8&74.6 \\ \hline
			80.0&81.3&73.1&74.5&72.7&76.9&72.7 \\ \hline
			82.5&83.9&76.7&71.9&69.0&76.0&71.6 \\ \hline
			85&86.5&80.9&68.4&66.6&74.6&71.6 \\ \hline
		\end{tabular}
	\end{table} 
	The positive class, diabetics, is the worst classified, with an average True Positive Rate (TPR) of $60.8$ over the ten training subsets and $56.4$ over the ten test subsets. The negative class, non-diabetics, presents an average True Negative Rate (TNR) of $90.5$ and $87.8$ over the ten training and test subsets, respectively. In this setting, it is preferable to better classify diabetic patients, since diagnosing a diabetic as non-diabetic is more critical (in terms of missing medical treatment) than the other way around.
	
	In this regard, a performance constraint over the positive class has been added to the ORCT for several values of the threshold $\rho_{+}$. We have considered a grid of values for $\rho_{+}$ varying from $62.5$, a slightly higher value than the training TPR obtained with $\rho_{+}=0$, to $85.0$ in steps of $2.5$ units.
	Note that constraints on expected performance \eqref{performanceconstraints} are imposed over the training sample, so they might not be satisfied on an independent sample.
	Indeed, results in Table \ref{Results with constraints on expected performance over the positive class in the Pima-indians-diabetes data set.} show how these thresholds are fulfilled in the training subsets but this is not necessarily the case in the test subsets; even so, we observe that the test TPR increases with $\rho_{+}$. \figurename{ \ref{tprtraintest}} supports this observation, in which the $\text{TPR}_{\text{train}}$ and the $\text{TPR}_{\text{test}}$ are depicted as a function of the imposed $\text{TPR}\,\,(\rho_{+})$. There, we can see that there exists a good linear fit. In fact, the regression models' coefficients and their corresponding coefficients of determination are the following:
	\begin{equation*}
	\text{TPR}_{\text{train}} = -0.36+1.02\rho_{+}, R^2 = 0.9993,
	\end{equation*}
	\begin{equation*}
	\text{TPR}_{\text{test}}= -0.65+0.94\rho_{+}, R^2 = 0.9754.
	\end{equation*} A clear overfitting, almost independent of the threshold imposed, is also detected; but it is possible to determine the required imposed TPR in order to obtain a successful $\text{TPR}_\text{test}$. There is a price to pay for achieving such high TPRs: the TNRs decrease as we demand larger thresholds, see \figurename{ \ref{tnrtest}}.
	
	\begin{figure}
		\centering
		\includegraphics[scale=0.5]{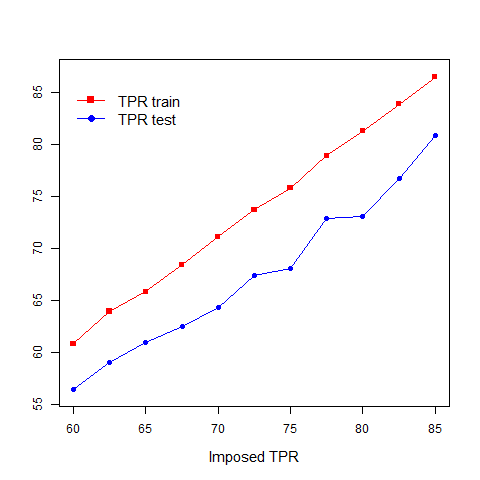}
		\caption{$\text{TPR}_{\text{train}}$ and $\text{TPR}_{\text{test}}$ drawn as a function of the imposed $\text{TPR}\,\,\bm{(\rho_{+})}$ for Pima-indians-diabetes data set.}.
		\label{tprtraintest}
	\end{figure}
	
	\begin{figure}
		\centering
		\includegraphics[scale=0.5]{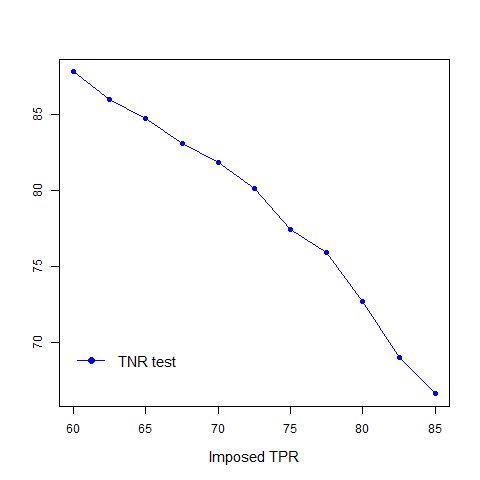}
		\caption{$\text{TNR}_{\text{test}}$ depicted as a function of the imposed $\text{TPR}\,\,\bm{(\rho_{+})}$ for Pima-indians-diabetes data set.}
		\label{tnrtest}
	\end{figure}
	
		\section{Conclusions and future research}
		\label{Conclusions and future research}
		
		Several papers have been proposed in recent years to build classification trees in which the greedy suboptimal construction approach is replaced by solving an optimization problem, usually in integer variables. These procedures, while successful against CART, are very time consuming and can only address problems of moderate size. In this paper we have proposed a new optimization-based approach to build classification trees. By replacing the binary decisions with randomized decisions, the resulting optimization problem is smooth and only contains continuous variables, allowing one to use gradient information. The computational experience reported shows that, with low running times, we outperform recent benchmarks, getting closer to and some times better than the performance of Random Forests. Moreover, we can model cost-sensitive constraints, having a direct and effective control on the accuracy of critical classes.
		
		Several extensions of our approach are promising and deserve further investigation. First, it is known that bagging trees (i.e., using Random Forests instead of decision trees) tends to enhance the accuracy. An appropriate bagging scheme for our approach is a nontrivial design question. Second, if the user wants to improve both accuracy and computing times using deeper trees, at the expense of interpretability, then one can develop a local-search procedure to embed our algorithm, as done successfully in \cite{dunn2018optimal} for the integer programming approach in \cite{bertsimas2017optimal}. Third, it is difficult in decision trees to control the number of predictor variables used. Making our approach sparse by means of an $\ell_1$ regularisation, i.e., by using a lasso-type objective, is also an interesting research question.

\section*{Acknowledgements}

This research has been financed in part by research projects EC H2020 MSCA RISE NeEDS (Grant agreement ID: 822214),  FQM-329 and P18-FR-2369 (Junta de Andaluc\'{\i}a), and PID2019-110886RB-I00 (Ministerio de Ciencia, Innovación y Universidades, Spain). This support is gratefully acknowledged.

\bibliography{biblio}

\end{document}